\documentclass[final,3p,times,twocolumn]{elsarticle}
\usepackage[utf8]{inputenc}
\usepackage{booktabs}
\usepackage{tabulary}
\usepackage{longtable}
\usepackage{ulem}
\usepackage{url}

\title{A Framework for Constructing Machine Learning Models with Feature Set Optimisation for Evapotranspiration Partitioning}

\author[1]{Adam Stapleton} \author[2]{Elke Eichelmann}
\author[3]{Mark Roantree}
\date{July 2021}

\address[1]{School of Computing, Dublin City University, Dublin 9, Ireland\\}
\address[2]{School of Biology and Environmental Science, University College Dublin, Dublin 4, Ireland\\}
\address[3]{Insight Centre for Data Analytics, Dublin City University, Dublin 9, Ireland}

\DeclareGraphicsExtensions{.pdf,.png,.jpg,.jpeg,.JPG,.PNG}

\usepackage{gensymb}

\usepackage{amsmath}
\usepackage[ruled,vlined]{algorithm2e} 
\usepackage{xspace}
\usepackage{multirow}
\usepackage{graphicx}
\usepackage{subcaption}
\usepackage{float}
\usepackage{tabularx}

\begin{document}

  
\begin{abstract}
A deeper understanding of the drivers of evapotranspiration and the modelling of its constituent parts (evaporation and transpiration) could be of significant importance to the monitoring and management of water resources globally over the coming decades. In this work, we developed a framework to identify the best performing machine learning algorithm from a candidate set, select optimal predictive features as well as ranking features in terms of their importance to predictive accuracy. 
Our experiments used 3 separate feature sets across 4 wetland sites as input into 8 candidate machine learning algorithms, providing 96 sets of experimental configurations. Given this high number of parameters, our results show strong evidence that there is no singularly optimal machine learning algorithm or feature set across all of the wetland sites studied despite their similarities.
A key finding discovered when examining feature importance is that methane flux, a feature whose relationship with evapotranspiration is not generally examined, may contribute to further biophysical process understanding.
\end{abstract}

\maketitle



\section{Introduction}
\label{sec:Introduction}

Evapotranspiration (\textit{ET}) is the process by which water is exchanged between the biosphere and the atmosphere. Better understanding of ET processes and their drivers in various environments is important for the entire terrestrial hydrological cycle that governs the transport and recycling of the water that supports, for example, our fresh water supplies \cite{oki2006global} \cite{zeng2018impact}. Observations of the Earth's atmosphere and biosphere over the last number of decades have indicated an intensifying hydrological cycle \cite{brutsaert1998hydrologic} \cite{pascolini202110} and an increase in the number of people living in water stressed areas \cite{oki2006global}. Modelling efforts over this period have shown disagreements, with evidence indicating a decline in global terrestrial \textit{ET} caused by a reduction in available moisture supply \cite{jung2010recent} and more recently, indication of an increase in global terrestrial ET due to increasing land temperature \cite{pascolini202110}. \textit{ET} is a process composed of two main parts: Evaporation (\textit{E}), the physical process, and Transpiration (\textit{T}), a biologically modulated process that occurs through the stomata of plants. A better understanding of the drivers of ET and the modelling of each of its constituent parts will be of significant importance to the monitoring and management of water resources globally over the coming decades. ET research contributes to many important components of global climate modelling including cloud formation (of relevance due to their role in the absorption and reflection of solar radiation and the transfer of energy between environments) and moisture availability \cite{gerken2018surface} \cite{green2017regionally} \cite{pielke1998interactions} \cite{schlesinger2014transpiration} \cite{trenberth2009earth}. The partitioning of ET into its constituents is vital in reducing the associated uncertainty in climate land surface models and satellite remote sensing projects such as ECOSTRESS \cite{fisher2020ecostress} as current models are validated on combined ET data only \cite{stoy2019reviews}.
The usage of machine learning (ML) in the domain of biosphere-atmosphere exchange has seen an increase in recent years with the availability of large, open source Eddy Covariance (EC) data sets such as FLUXNET \cite{baldocchi2001fluxnet} and AmeriFlux \cite{novick2018ameriflux} enabling more data intensive approaches.
Applications of ML in the domain of biosphere-atmosphere exchange have mostly focused on gap-filling of EC data \cite{irvin2021gap} but some success has been achieved in the application of ML techniques to partitioning of gas fluxes \cite{tramontana2020partitioning}, prediction of fluxes \cite{tramontana2016predicting}, spatial interpolation \cite{lin2002comparison}, and upscaling of EC data \cite{jung2009towards, bodesheim2018upscaled}.
As the EC method measures total water flux, the goal of partitioning in this work is to determine the individual contributions of \textit{E} and \textit{T} to the net flux. 
\section{Background}
\label{sec:Background}

\subsection{Data}
\label{sec:Data}

The data utilised in this work are obtained using the Eddy Covariance (EC) method \cite{aubinet2012eddy} from measurement towers across four wetland sites in the Sacramento-San Joaquin river delta in Northern California: West Pond (WP) \cite{valach2021WPdata}, East End (EE) \cite{eichelmann2021EEdata}, Mayberry Farms (MB) \cite{matthes2021MBdata}, and Sherman Island (SW)\cite{shortt2021SWdata}. This method ascertains the flux of trace gases by measuring the covariance between fluctuations in vertical wind velocity and the mixing ratio of the gas in question.  The data from all sites are available under an open-source license as part of the AmeriFlux network and can be accessed through the AmeriFlux data sharing platform \cite{novick2018ameriflux} \cite{ameriflux2021data}.
The sites have been described in detail in other publications (\cite{detto2010scaling}, \cite{hatala2012greenhouse}, \cite{knox2015agricultural}, \cite{eichelmann2018effect}) and the reader is referred to these works for a more complete description. 
The four sites are all freshwater marsh wetlands that have been constructed by the Department of Water Resources to manage soil subsidence in the area. The observation period for each site differs in length with approximately 10 years of data for MB (October 2010 to October 2020), 8 for WP (July 2012 to September 2020), 7 for EE (November 2013 to September 2020) and 4 for SW (May 2016 to April 2020). All sites, with the exception of WP, underwent flooding within the measurement period. WP however was established in 1998, making it the longest standing of the four sites. The initial flooding period is of note as it provides a period in which vegetation has not yet been established and thus, it can be assumed that \textit{T} is negligible during this period. The vegetation cover (within the EC footprint at the latest measurement in 2018 \cite{valach2021productive}) varies between the sites: 97\% at WP, 64\% at MB, 96\% at EE and 45\% at SW. The lower vegetation cover at SW can be explained by the fact that it is the newest wetland to be established, constructed in 2016. The dominant vegetation species at all sites are tule (\textit{Schoenoplectus acutus}) and cattail (\textit{Typha} spp.) \cite{o2015hybrid}.
Continuous fluxes of water vapour and other trace gases were measured using the EC method.
In addition to the EC data, micro-meteorological and environmental data were also obtained for each of the sites including the following variables with a known relationship with ET; air temperature (\textit{TA}); water temperature (\textit{TW}); soil temperature (\textit{TS}); relative humidity (\textit{RH}); atmospheric pressure (\textit{AP}); net radiation (\textit{RNET}); water table depth (\textit{WT}); vapor pressure deficit (\textit{VPD}); sensible heat exchange (\textit{H}); friction velocity (\textit{u*}); vegetation greenness index from camera data (\textit{GCC}) and the target variable water flux (labelled total ET or \textit{wq}). The data frequency is at 30 minute intervals and where the data were recorded at higher frequencies, the mean was computed for that 30 minute period. Pre-processing of the data to remove spikes, filter for instrument malfunctioning and gap-filling procedure for certain data has been described in detail in \cite{eichelmann2018effect}.
For EC flux features with missing data, a Neural Network (NN) procedure was used for imputation \cite{baldocchi2015does} \cite{knox2015agricultural} and is detailed in our previous work \cite{Eichelmann2021ET}. Meteorological variables were imputed using data from nearby weather stations where data were available. For any remaining features with missing data linear interpolation or mean imputation was used.

\subsection{Machine Learning Algorithms}

In this paper, a variety of supervised ML algorithms were utilised and the resulting models compared for performance on the prediction task described in Section \ref{sec:novel}. The algorithms tested can be broadly grouped into the following 3 categories: parametric regressors, non-parametric regressors and ensembles \cite{geron2019hands}. The scikit-learn library \cite{scikit-learn} was used for model building in addition to the XGBoost \cite{chen2016xgboost} and LightGBM \cite{ke2017lightgbm} libraries. 

Parametric models, such as linear and ridge regression models or NNs, produce a predictive function by assuming a model with a fixed number of parameters and improving the performance of the model by adjusting the weights of the parameters until a minimum loss is obtained. Due to their simplicity, linear parametric models are extremely fast in training and prediction though may suffer from underfitting if the true distribution of the data is more complex.

Non-parametric models, such as \textit{K}-nearest neighbours (KNN), decision trees (DT) and Support Vector Machines (SVM) make little or no assumptions about the predictive function in advance and seek to learn both the functional form and the function's parameter values from the data. Non-parametric models produce a more flexible predictive function, thereby allowing them to better model more complex distributions. This increase in complexity can lead to overfitting and an increase in both training time and the volume of data required to fit more complex models. 

Ensemble methods such as Gradient Boosting, XGBoost and LightGBM, combine the predictions of many simple models, referred to as weak learners or base learners, to produce a predictive model. Base learners can be trained in parallel and combined using methods such as bagging or stacking,  or sequentially using methods such as
boosting \cite{geron2019hands}. For all ensemble methods tested, the base learners were DT. Ensemble methods are generally less prone to overfitting while still retaining sufficient complexity to arrive at a reasonable approximation of the underlying distribution.


%
%
%
%
\section{Framework Methodology}
\label{sec:Methodology}
\begin{figure*}[h!]
\centering
\includegraphics[scale = 0.5]{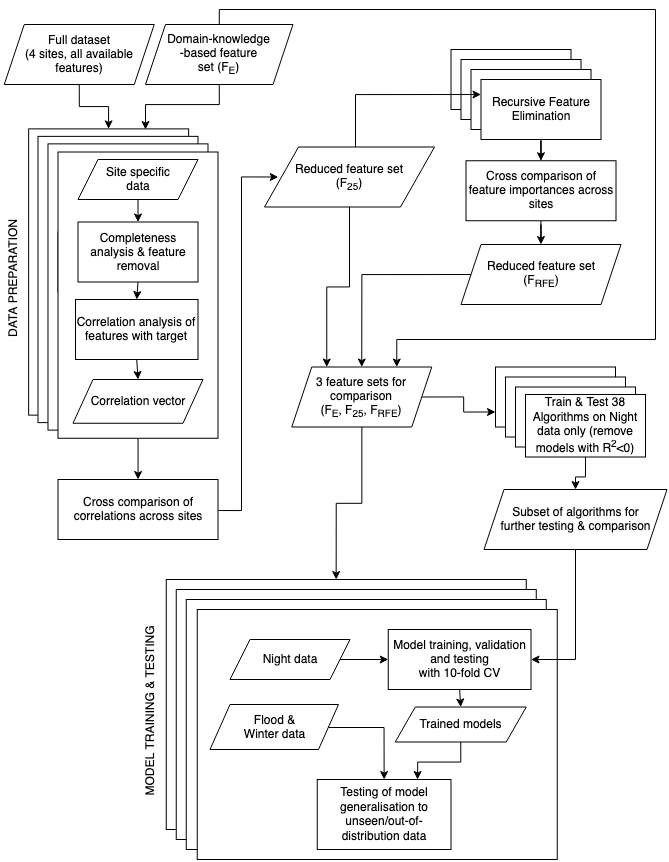}
\caption{Process flow for the entire framework. Each rectangle represents a computational process and each parallelogram an input/output to that process. Where 4 layers are seen in a set of processes, this indicates that the processes within that set were carried out separately on the data for each individual site. Where the process is not layered or within a set of processes, this indicates it was carried out across all sites in combination.}
\label{fig:Flow}
\end{figure*}

This section begins with a description of our approach to ET Flux Partitioning. The proposed framework is then described as the following series of methodological steps: data preparation, feature selection, construction of baseline models, final model training, evaluation and comparison. These steps are represented as a flow in Figure \ref{fig:Flow}

Each set of experiments were repeated across the four wetland sites for each set of algorithms. In order to provide fair comparison between each of the algorithms either the same data were used in training and evaluation or a cross-validation procedure was applied in training and validation. 

\subsection{Evapotranspiration Partitioning}
\label{sec:novel}
A novel, data-driven method to ET partitioning (drawing from previous work on  carbon dioxide flux partitioning \cite{tramontana2020partitioning}) was presented previously in \cite{Eichelmann2021ET}, with a brief outline here. Given the difficulty in establishing ground-truth data for the component contributions of \textit{E} and \textit{T} to overall $ET$, a number of assumptions are used to establish periods during which \textit{T} can be assumed to be negligible and therefore, taken to be 0 in calculations. During the night, plant stomata are assumed to be closed and therefore not transpiring (confirmed with leaf level measurements at these sites). Utilising this assumption, the night-time data (Night) are used to train models to predict \textit{E}, which can then be subtracted from total $ET$ to give predicted value for \textit{T}.
Explicitly the relationship between \textit{E} and \textit{T} can be expressed using Equation \ref{ET:basic}.

\begin{equation}\label{ET:basic}
\begin{split}
    ET = T + E \\
    T_{Night} \simeq 0 \\
    ET_{Night} = E
\end{split}   
\end{equation}

As there are no measured ground truth data for the individual components \textit{E} and \textit{T}, assumptions about \textit{T} during other periods of the year are used to determine two testing sets, namely the data from the initial flooding period (Flood) and from the winter senescent months (Winter).  
During the initial flooding period of each of the wetland sites, vegetation had not yet been established and therefore, \textit{T} was not occurring. During the winter months the vegetation are observed to be senescent and here again, \textit{T} is negligible. An additional set of core assumptions are used in determining the timing of the onset and duration of these periods. The zenith angle of the sun being greater than 90\degree~is used to determine the night-time periods. Visual determination of the level of vegetation from camera observation of the sites is used to determine the onset of vegetation after the initial flooding period, also referred to as "greenup". Lastly, the months of December, January and February are taken as the senescent periods. Limitations of these assumptions is discussed in Section \ref{sec:Limitations}.

\subsection{Data Preparation}
\label{sec:dataprep}

%
In order to reduce the dimensionality (number of features) of the data to be computationally tractable for model building the following approach to feature selection was undertaken; first reduce the candidate number of features using correlation and completeness analyses, then explore possible feature sets of different sizes and in different combinations using Recursive Feature Elimination. This process is described fully in Section \ref{sec:RFE}. 

As in \cite{Eichelmann2021ET}, domain knowledge was used to inform the selection of features that have a known relationship with water flux. These were \textit{VPD, GCC, u*, TA, RNET, WT, H} and ecosystem respiration estimated from an exponential relationship between night-time carbon flux and temperature as performed in \cite{reichstein2005separation} (\textit{ER\textsubscript{Reichstein}}). In addition, three time features were added: \textit{year}, \textit{month} and \textit{day of the year (DOY)}. This forms our first feature set for testing, denoted by the identifier \textit{F\textsubscript{E}}. 
All features with a completeness less than 80\% (i.e. missing greater than 20\% of the data) for the measurement period were discarded. Soil and water temperature measurements taken at various depths showed low levels of completeness and the depths at which measurements were taken was not consistent across sites. In order to obtain a usable feature for soil and water temperature that is comparable between sites, the measurements were consolidated by computing the mean of the sensor values across all depths to create two new features, \textit{TS (mean)} and \textit{TW (mean)}. 
For the remaining features mean imputation was used to replace the remaining missing data and a correlation analysis was undertaken to extract the most likely useful features. Of the 50 most highly correlated features at each site, the 25 features that were common across all sites in that subset of 50 were selected and added to the \textit{F\textsubscript{E}} feature set. The resultant feature set was labelled \textit{F\textsubscript{25}}. This approach was taken as it is hypothesised by the authors that a correlation with the target feature that is common across multiple sites will be more likely to have an underlying physical causal relationship.

\subsection{Model Comparison}
\label{sec:model_comparison}
In order to to test and compare a suitably diverse set of algorithms for model building, initial testing examined 38 algorithms from the scikit-learn library \cite{scikit-learn} alongside two additional ensemble algorithms; LightGBM \cite{ke2017lightgbm} and XGBoost \cite{chen2016xgboost}. The models are compared in order to ascertain which algorithm will be most suited to the model building task, including but not limited to an improvement in model performance in terms of fitting well to the training data, generalising well to unseen data and computational cost.
Any models whose predictions had a negative coefficient of determination (R\textsuperscript{2}) with the target at any site were immediately discarded. From the remaining models, a subset of the best performing models (or simplest model in the case of equal model performance) were selected across 3 different categories of models; parametric, non-parametric and ensemble. The inclusion of different categories of ML algorithms is undertaken to prevent a loss in diversity from the initial set of algorithms tested. The models selected for final testing and comparison were as follows: linear regression,  ridge regression, KNN, DT, Gradient Boosting Decision Trees, LightGBM and XGBoost, with default hyper-parameters for all algorithms. A 10-fold cross validation was employed in training and testing across the Night, Winter and Flood datasets.


\subsection{Recursive Feature Elimination}
\label{sec:RFE}
In order to identify a feature set that contains maximal information with the minimum number of features a recursive feature elimination (RFE) method is used. A lower number of features is desired so as to reduce model complexity and as a result, reduce the chance of overfitting and to combat the so-called "curse of dimensionality" \cite{han2011data} whereby an increase in the number of features leads to a lower number of samples per unit volume of the feature space. In our method, a LightGBM model is trained on Night data with the features from the F\textsubscript{25} feature set for each of the four sites. Each model is then used to obtain a metric for the relative importance of each feature at that site. The metric that is used to measure feature importance is the sum of the gains in model performance, as measured by reduction in Root Mean Squared Error (RMSE), of all branches of base learner DT using that feature. The least important feature is then removed from the feature set, a new model is then trained on the resulting, smaller feature set and the process is repeated until no features remain. Cross validation is applied at each iteration and the mean of the model performance metrics are recorded. The optimal features for each site are determined to be the feature set that preceded a 1\% decrease in R\textsuperscript{2} for the hold-out Night data as the number of features is iteratively decreased. The feature sets obtained for each site are then compared for commonalities and those features that were in the optimal feature set for only one site are discarded and the remaining feature set is labelled \textit{F\textsubscript{RFE}}. We hypothesise that this feature set approaches the minimum number of features needed to capture all information needed to model \textit{E} and \textit{T} from the available data.

\subsection{Evaluation}
\label{sec:eval}

In order to select the most appropriate set of metrics for evaluating model performance, the nature of the data must be taken into consideration. In contrast to a conventional supervised learning problem where the ground-truth data were obtained under known conditions, the ground-truth data for the experiments in this study are based on an assumption about approximate levels of \textit{T} occurring under different conditions. During the night-time, winter and initial flooding periods, the assumptions governing negligible \textit{T} are slightly different. Therefore, we expect that some common metrics for the evaluation of a regressor (e.g. RMSE and mean average error) may lead to difficulty in comparing model performance across test sets as the level of actual \textit{T} occurring may vary and be non-negligible in some cases. This may lead to increases in measures of predictive error that are not attributable to poor predictive performance but rather to deviations in the data caused by a confounding variable that is not present in the training data (namely \textit{T} arising in total measured ET where the model assumes that the total measured ET should be measuring \textit{E} only). 

In addition, each of the Night, Flood and Winter datasets have different data distributions \cite{Eichelmann2021ET} and it is the performance of the models on data whose values lie outside the range of the training data (referred to as unseen data) that must be evaluated.

Therefore, a metric that determines how closely the variations in predictions of \textit{E} follow the variations in total measured \textit{ET} across all test sets is required. For this reason, the metrics chosen for evaluation are R\textsuperscript{2}, Adjusted R\textsuperscript{2} (R\textsuperscript{2}\textsubscript{Adj}) and slope of line of best fit between ground truth and predictions (\textit{m}). R\textsuperscript{2}\textsubscript{Adj} enables comparison between feature sets as this metric adjusts for the number of features used in order to account for the often spurious increase in R\textsuperscript{2} when additional features are added to a model.

\begin{equation}\label{eqn:adjR2}
    R^{2}_{Adj} = 1 - \frac{(1-R^2)\times(p-1)}{p-q-1}
\end{equation}

Equation \ref{eqn:adjR2} describes Adjusted R\textsuperscript{2} where R\textsuperscript{2} is the R\textsuperscript{2} of the model, $p$ is the number of samples and $q$ is the number of features.

Slope is chosen in order to validate one of the biophysical constraints of any partitioning model, namely that the slope of the line of best fit between the predicted \textit{E} and the ground truth (total \textit{ET}) never exceeds 1 for any of the data. A slope greater than 1 would indicate that \textit{E} had exceeded net ET which would lead to negative \textit{T}, violating the biophysical constraint that negative \textit{T} cannot occur.

These metrics are obtained for a subset of the Night data that is not used in training or validation (referred to as the hold-out or test set) as well as on the entire Flood and Winter data.

%
%
%
%

\section{Results \& Discussion}
\label{sec:Results}


In this section experimental results are presented where, for all four wetland sites, identical feature sets and experimental configurations were used. The results are reported on a per site basis as the goal is to compare how each of the models generalise to unseen data for the same site they were trained on. All results are the mean values of the metric across 10 cross-validation folds. 

%
%
\begin{figure*}[h]
\centering
\includegraphics[scale = 0.48]{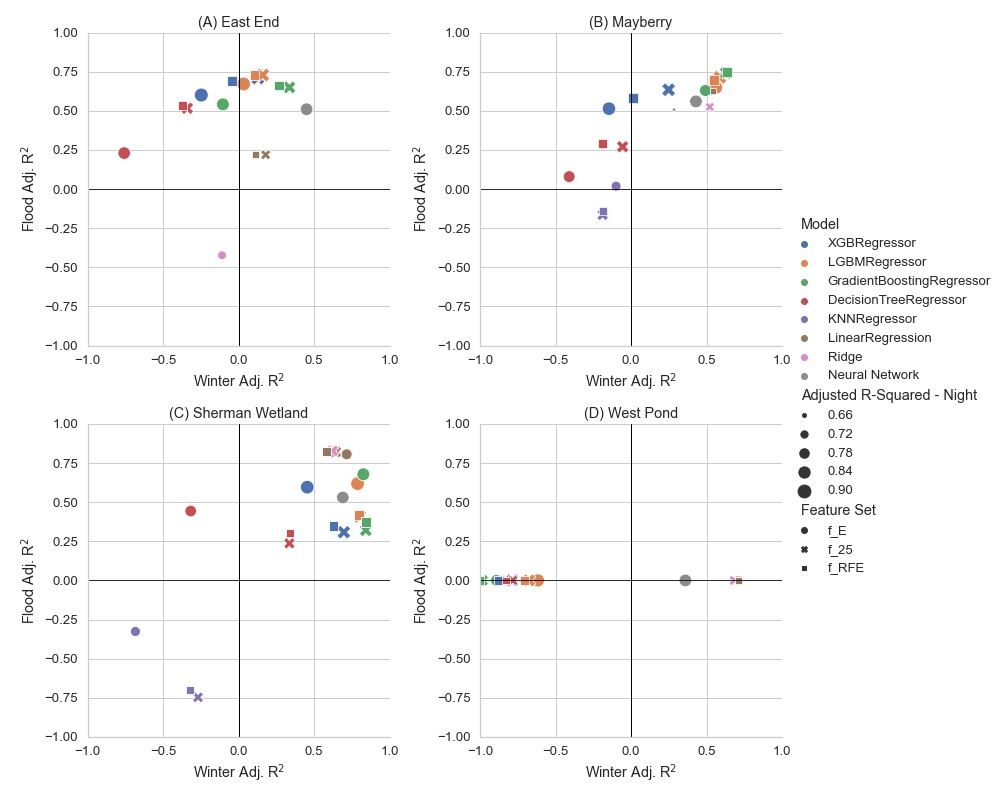}
\caption{Results of model comparison for the four sites being studied.}
\label{fig:Model Comparison Results}
\end{figure*}

\begin{figure*}[h]
\centering
\makebox[0pt]{%
    \includegraphics[scale = 0.38]{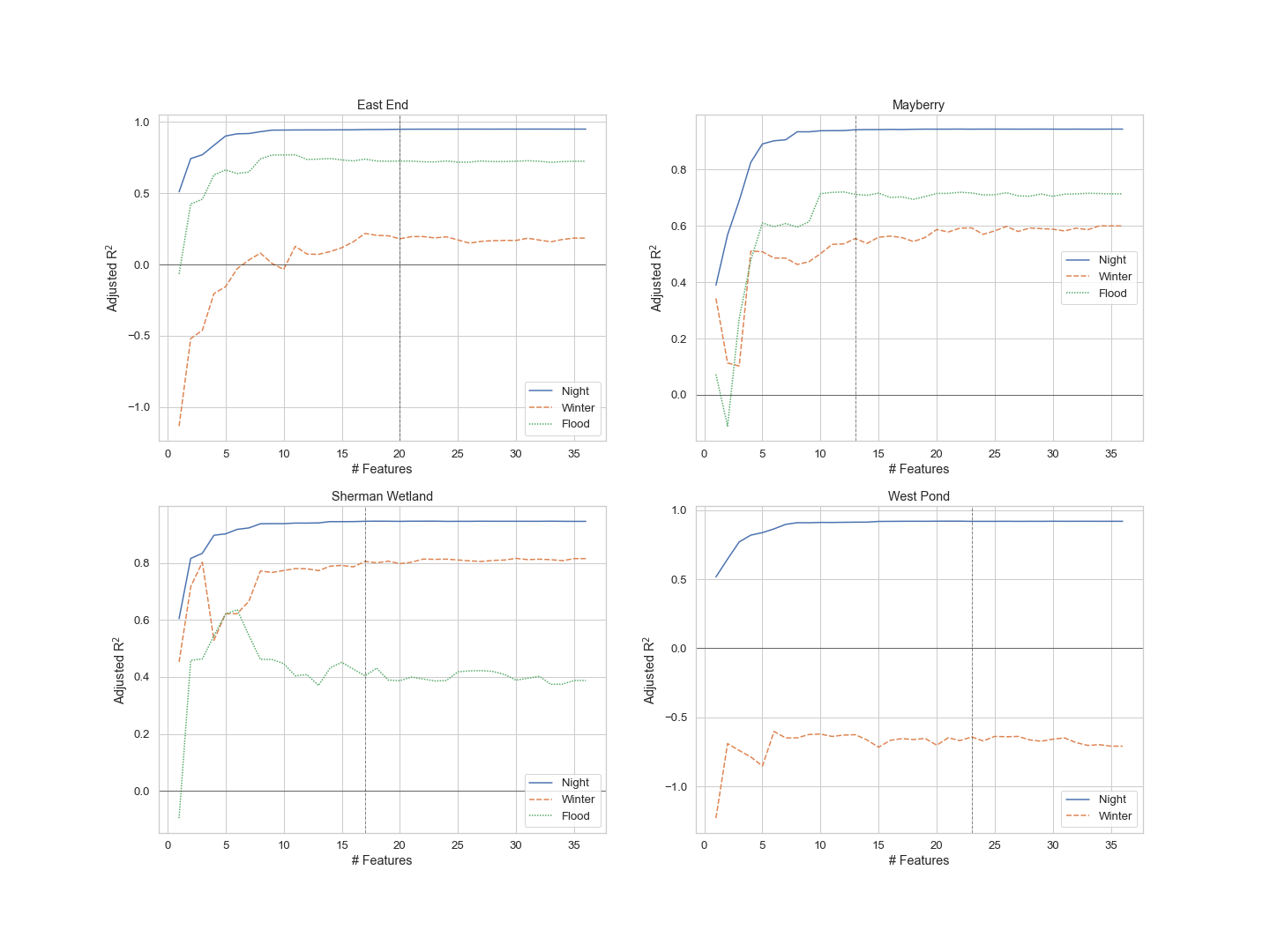}}
\caption{Results of the RFE process for each of the 4 sites tested }
\label{fig:RFE Results}
\end{figure*}
\subsection{Model Comparison Results}
\label{sec:Model Comparison Results}

Figure \ref{fig:Model Comparison Results} shows the R\textsuperscript{2}\textsubscript{Adj} values for all sites, algorithms and feature sets tested for the Night, Winter and Flood data. The x-axis plots the R\textsuperscript{2}\textsubscript{Adj} values for predictions on the Winter data. The y-axis plots the Adjusted R\textsuperscript{2} values for predictions on the Flood data, testing the ability of the models to generalise to unseen data. The colour of the marker indicates the algorithm used in model building and the shape of the marker indicates the feature set being tested.

The size of the marker indicates the R\textsuperscript{2}\textsubscript{Adj} values for predictions on the hold-out Night data, demonstrating how well the models perform on data that is identically distributed to the training data. Therefore, the best performing models are those with the largest markers that are closest to the upper right corner of the graph. The x- and y-axis lines along the origin are displayed to allow for ease of identification of those models that fail to generalise well (i.e. models with $R\textsuperscript{2}\textsubscript{Adj}<0$). As WP does not have Flood data, the results are displayed along the x-axis only.


Figure \ref{fig:Model Comparison Results} shows that an improvement in model performance was obtained on Night data as well as in generalising to Winter and Flood data over and above that of our previous results \cite{Eichelmann2021ET}, where the previous results are those that utilised a NN-based models and the F\textsubscript{E} feature set, indicated by a grey circular icon. 

In general, results show that addition of the extra 25 features from the correlation analysis gave some improvement in model performance across all model types when compared to the baseline feature set, \textit{F\textsubscript{E}}. In addition, it is seen that reduction in features from 36 to 20 in going from \textit{F\textsubscript{25}} to \textit{F\textsubscript{RFE}} either resulted in further incremental improvement for the best performing models, or did not drastically decrease model performance. 
Interestingly, all sites had more than one model which failed to generalise well to the Winter and Flood data. 
This observation is important as it may indicate that a site-specific approach to model building may be more favourable. The sites modelled in this paper are all biologically similar: all wetlands with the same species composition, same climate and similar management. As described in Section \ref{sec:Data} there are some known differences between the sites, such as the ratio of open water to vegetation cover and the utility of our framework may be best realised when building models that contain not only the general features relevant for modelling a particular ecosystem but also those features that are relevant for modelling that particular site. 

LightGBM based models performed well at all sites except WP with very low computational resource indicating that it would be an ideal choice of algorithm in most cases. Furthermore, its reduction in computational time, resource requirements and high predictive performance on Night, Flood and Winter data would suggest an ideal candidate model.


Many of the features are known to exhibit a high-level of non-linearity in their relationship with ET, particularly when considering the shift from night to day. For example, \textit{RNET} is approximately constant and negative at night while positive
This may go towards explaining why simpler models (such as linear parametric models or models that used less features) performed worse in many cases - the generated hyper-plane  may not have sufficient complexity to model the underlying relationship between the predictive features and the target feature.

It is also noted that most of the models failed to generalise well for the Winter data at WP, indicating that there may be particularities about this site that were not captured in the features or in the learned predictive function. This may be due to the fact that WP has differences in its composition to the other sites being the oldest of the 4 sites, as previously discussed in Section \ref{sec:Data}.



\subsection{RFE Results}
\label{sec:RFE Results}

Figure \ref{fig:RFE Results} displays the results of the RFE process with number of features on the x-axis and R\textsuperscript{2}\textsubscript{Adj} results on the y-axis. The iterations start on the right and move towards 0 as RFE iteratively decreases the number of features until only 1 feature remains for each of the sites and each of the test sets; Night, Winter and (where available) Flood.  A vertical line on each graph indicates the number of features selected, where the optimal feature set is determined to be the last feature set preceding a 1\% reduction in R\textsuperscript{2}\textsubscript{Adj}. 


\textit{F\textsubscript{RFE}}, the feature set generated by the RFE process, contains all features from  \textit{F\textsubscript{E}} except \textit{GCC} and \textit{TA}. These features may have resulted in less information gain than other features due to the large percentage of imputed data for each of these features, despite their known physical relationship with ET. It is also noted that the temperature information may already be captured sufficiently in the \textit{TW} or \textit{TS} variables.


It is evident that the number of features needed for an optimally performing model varies from site to site, indicating the difficulties in determining a universally optimal feature set. For example at SW we can see that a model with just 3 features generalises best on Winter data and generalises better than the feature set chosen by the RFE process for that site. However, we can also observe that a model with 6 features for that site generalises best on Flood data with a reduction in performance in generalising to Winter data. 



%
%

\subsection{Feature Importance}
\label{sec:Feature Importance}

\begin{table*}[]
\centering
\resizebox{\textwidth}{!}{%
\begin{tabular}{c|llllllll}
\multicolumn{1}{l|}{\textbf{Rank}} &
  \multicolumn{1}{l}{\textbf{F (EE)}} &
  \multicolumn{1}{l}{\textbf{I (EE)}} &
  \multicolumn{1}{l}{\textbf{F (SW)}} &
  \multicolumn{1}{l}{\textbf{I (SW)}} &
  \multicolumn{1}{l}{\textbf{F (MB)}} &
  \multicolumn{1}{l}{\textbf{I (MB)}} &
  \multicolumn{1}{l}{\textbf{F (WP)}} &
  \multicolumn{1}{l}{\textbf{I (WP)}} \\ \hline
\textbf{1}  & \textbf{TW\ (mean)}    & 0.289                & \textbf{u (mean)}       & 0.549                         & \textit{wm}\     & 0.180                & \textbf{H}\         & 0.440 \\
\textbf{2}  & \textbf{u (mean)}        & 0.191                & VPD        & 0.169                         & u*      & 0.170                & u*         & 0.186 \\
\textbf{3}  & \textit{c (mean)}        & 0.127                & u*      & 0.049                         & year       & 0.112                & RNET          & 0.097 \\
\textbf{4}  & RH          & 0.076                & H\      & 0.047                         & RH         & 0.112                & VPD           & 0.054 \\\cline{4-5}
\textbf{5}  & year        & 0.073                & TW\ (mean)   & 0.041                         & u (mean)       & 0.092                & RH            & 0.047 \\ \cline{8-9}
\textbf{6}  & VPD         & 0.052                & wm\     & 0.034                         & TW\ (mean)   & 0.082                & DOY           & 0.034 \\
\textbf{7}  & u*       & 0.049                & RH         & {0.026}                & \textit{c (mean)}       & 0.075                & year          & 0.022 \\\cline{2-3}
\textbf{8}  & H\       & 0.033                & DOY        & 0.022                         & DOY        & 0.060                & \sout{TA}            & 0.017 \\
\textbf{9}  & TS\ (mean)    & 0.023                & t (mean)       & 0.021                         & VPD        & 0.059                & uw            & 0.017 \\\cline{6-7}
\textbf{10} & DOY         & 0.022                & time       & {0.008}                & H\      & 0.020                & wm\        & 0.014 \\
\textbf{11} & uw          & 0.017                & uw         & 0.007                         & ER\textsubscript{linear} & 0.016                & \sout{uu}            & 0.013 \\
\textbf{12} & RNET        & 0.014                & ww         & 0.007                         & ww         & 0.012                & WD            & 0.010 \\
\textbf{13} & ww          & 0.007                & RNET       & 0.006                         & WD         & 0.010                & TS\ (mean)      & 0.009 \\
\textbf{14} & {WT} & 0.006                & year       & 0.004                         &            & \multicolumn{1}{l}{} & time          & 0.007 \\
\textbf{15} & \sout{stat\textsubscript{q}}     & 0.004                & TS\ (mean)   & {0.004}                &            & \multicolumn{1}{l}{} & TW\ (mean)      & 0.006 \\
\textbf{16} & wm\      & 0.004                & ER\textsubscript{linear} & 0.004                         &            & \multicolumn{1}{l}{} & \sout{sos}           & 0.004 \\
\textbf{17} & ze          & 0.004                & WD         & 0.002                         &            & \multicolumn{1}{l}{} & ER\textsubscript{linear}    & 0.004 \\
\textbf{18} & time        & 0.003                &            & \multicolumn{1}{l}{}          &            & \multicolumn{1}{l}{} & WT            & 0.004 \\
\textbf{19} & t (mean)        & 0.003                &            & \multicolumn{1}{l}{}          &            & \multicolumn{1}{l}{} & \sout{ts (mean)}         & 0.004 \\
\textbf{20} & WD          & 0.003                &            & \multicolumn{1}{l}{{}} &            & \multicolumn{1}{l}{} & u (mean)          & 0.003 \\
\textbf{21} &             & \multicolumn{1}{l}{} &            & \multicolumn{1}{l}{{}} &            & \multicolumn{1}{l}{} & \sout{$\rho$\textsubscript{mol}} & 0.003 \\
\textbf{22} &             & \multicolumn{1}{l}{} &            & \multicolumn{1}{l}{}          &            & \multicolumn{1}{l}{} & \sout{vv}            & 0.003 \\
\textbf{23} &             & \multicolumn{1}{l}{} &            & \multicolumn{1}{l}{}          &            & \multicolumn{1}{l}{} & ze            & 0.003 \\ \hline
\end{tabular}%
}
\caption{Feature importance ranked in order of importance for each site where the features obtained by the RFE process are denoted by \emph{F} followed by the site label and the relative importance of that feature at that site is given by \emph{I} followed by the site label. Features that were omitted from the final feature set (F\textsubscript{RFE}) are indicated by a strike-through, highly important features indicated in bold, features of interest in italics and the threshold for significant feature importance indicated by a horizontal line for each site.}
\label{tab:Feature-Importance}
\end{table*}
Table \ref{tab:Feature-Importance} lists the features selected using the RFE process for \textit{each} site. Feature (F) columns rank features by their importance while Importance (I) columns give the relative proportion of total gain in performance contributed by that feature, normalised to sum to 1.
The features that were not included in F\textsubscript{RFE} are denoted by a strike-through. A full list of the features tested and their descriptions can be found in the supplementary material. 


As noted in Section \ref{sec:RFE Results}, there is an overlap with the features previously selected using domain knowledge and many of the new features selected relate to processes that are known mediators of \textit{E} and \textit{T}. Our previous work highlighted the importance of \textit{VPD} and \textit{u*} as they both relate to energy transport  \cite{Eichelmann2021ET} which affects \textit{E} as it is a form of latent energy. \textit{VPD} is a measure of dryness of the air which increases transport of water across this gradient from high to low moisture and \textit{u*} is a measure of turbulence  which also increases the transport of energy away from the surface. 


Most of the features idetified by our framework can be grouped into the following relationships with \textit{E}: the energy available for evaporation (\textit{TW}, \textit{H}, \textit{RNET}, \textit{TS}, \textit{tbar}), the moisture gradient driving \textit{E} (\textit{RH}, \textit{VPD} as well as \textit{WT} to a lesser degree)), the turbulent processes transporting water vapor away from the surface (\textit{u*},\textit{u (mean)}, \textit{uw}, \textit{ww}, \textit{WD} and \textit{stat\textsubscript{q}}) and
the temporal patterns of ET (\textit{year}, \textit{DOY}, \textit{time}, \textit{ze}). 
For description of variable labels please refer to Table \ref{tab:feature_meanings} in the supplementary material. 

If a threshold of 0.2 is set for highly important and 0.05 ($\pm$10\%) for significantly important, an examination of table \ref{tab:Feature-Importance} indicates 4 features (highlighted in bold) as being of high importance in accurately predicting ET: \textit{TW(mean)} and \textit{u(mean)} at the EE site; \textit{u(mean)} at the SW site; \textit{H} at the WP site; and no feature (meaning a more even distribution of importance across features) at the MB site. If the features that are deemed to be highly or significantly important are examined it is observed that EE has 7, SW has 4, MB has 9, and WP has 5 features. Effectively this means that the majority of the predictive performance is attributable to these features. Two variables (highlighted in italics) which were unexpectedly ranked as important were carbon dioxide concentration (\textit{c (mean)}) at EE and MB and methane flux(\textit{wm}) ranked as high importance at MB. We hypothesis that the relevance of the \textit{c (mean)} may be due to its connection to microbial activity via soil respiration wherein carbon dioxide and water are transported in the same way. 
The connection with \textit{wm} is not as clear as there are multiple pathways through which methane can be released; diffusion, ebullition, and plant mediated transport. The fact that \textit{wm} appears as an important predictive feature for \textit{E} could indicate that there is mostly diffusive transport occurring which would follow the same physical processes as evaporation. 

Identifying new features may reveal previously unknown connections between components of the system for further study with the potential to improve understanding of the underlying biophysical processes. This process is significantly enabled by our objective and data-driven framework.

%
%




\subsection{Limitations}
\label{sec:Limitations}


A large percentage of the data for the target has been imputed for all sites and additionally a small percentage of features were imputed with a variety of methods being used for imputation. Building models that use this data carry the errors and limitations of the imputation methods and may introduce noise to the data, particularly where linear or mean imputation was used. Assumptions around the onset of the different periods where \textit{T} is considered to be negligible may also lead to the introduction of noise to the target feature where \textit{T} could be low but non-negligible.


%
%
%
%
\section{Conclusions}
\label{sec:Conclusion}
In this work a new framework by which climate scientists can test the efficacy of multiple ML algorithms and identify the suitable predictive features from a high-dimensional candidate set has been presented. The result is a ranking of the candidate algorithms, a generally optimal feature set and an understanding as to how features contribute to model performance (predictive accuracy). For validation, micro-meteorological datasets were used and our framework to produce a model with an optimal balance between complexity and model performance. The framework adopts an  objective (i.e. without usage of domain knowledge) view of feature selection and demonstrated an improvement on the baseline \cite{Eichelmann2021ET} which used a subjective approach to feature selection.

Algorithm ranking identified that a LightGBM model would likely perform well on this task at other sites, generalising well to unseen data and requiring low computational resources. However this was not a universal result, with simpler linear parametric models performing best at WP, indicating that there are key differences between the sites that necessitate an approach that tailors the models to individual sites.

The RFE process identified new features from the data that improved model performance. The use of information gain as a metric to iteratively remove features also allows for a direct comparison as to which features were most important at each site, providing the basis for further work, either in transferring these learnings to new sites or refining the models for these sites. 

The examination of feature importance highlighted an obscure biophysical link in the case of carbon dioxide concentration and methane flux which improve our understanding of the physical and biological processes involved.

We conclude that this method provides new evidence of the contribution of ML to ET partitioning. The independence of the framework from explicit domain knowledge indicates that this approach may be domain agnostic, meaning that this method may have applications on other datasets, either for different EC flux sites or on entirely unrelated data. 

\textbf{Acknowledgement.}
  This work was funded by Science Foundation Ireland through the SFI Centre for Research Training in Machine Learning (18/CRT/6183) and by SFI Grant Number SFI/12/RC/2289\_P2, co-funded by the European Regional Development Fund. Funding for the AmeriFlux data portal was provided by the U.S. Department of Energy Office of Science. The authors would like to thank the site PI Dennis D. Baldocchi for enabling access to the extensive and high quality dataset without which this work would not have been possible. We would also like to acknowledge the technicians Joseph Verfaillie and Daphne Szutu and the many Berkeley Biometeorology Lab postdocs, PhD students and field assistants for their long years of work maintaining the AmeriFlux field sites and collecting and processing the data used in this study.

\bibliographystyle{agsm} 

\section{Appendix}
\label{sec:Appendix}

\begin{table*}[]
\centering
\resizebox{\textwidth}{!}{%
\begin{tabular}{l|l}
\toprule \hline
\multicolumn{1}{c|}{\textbf{F\textsubscript{25}}} & \textbf{Meaning}                                               \\
\hline \bottomrule
DOY                                & Day of Year                                                        \\
GCC                                & Vegetation greenness index from camera data                        \\
H                                  & Sensible heat exchange                                             \\
PA                            & Air Pressure                                                       \\
RH                                 & Relative Humidity                                                  \\
RNET                               & Net incoming and outgoing shortwave and longwave radiation; net radiation \\
TA                                 & Air temperature (from thermometer)                                 \\
TS (mean)                          & Mean soil temperature                                              \\
TW (mean)                          & Mean water temperature                                             \\
VPD                                & Vapor Pressure Deficit                                             \\
WD                                 & Wind Direction                                                     \\
WT                                 & Water Table Depth                                                  \\
c (mean)                           & Mean carbon dioxide concentration                                             \\
ER\textsubscript{Reichstein}                    & Ecosystem Respiration derived from Reichstein method               \\
ER\textsubscript{linear}                         & Ecosystem Respiration derived from Linear method                   \\
m (mean)                           & Mean methane concentration                                             \\
rho\textsubscript{mol}                           & Molar air density                                                  \\
sos                                & Speed of Sound                                                     \\
stat\textsubscript{m}                            & Degree of instationarity for mean methane concentration            \\
stat\textsubscript{q}                            & Degree of instationarity for mean water concentration              \\
stat\textsubscript{wc}                           & Degree of instationarity for wc covariance                         \\
stat\textsubscript{wm}                           & Degree of instationarity for wm covariance                         \\
stat\textsubscript{wt}                           & Degree of instationarity for wt covariance                         \\
t (mean)                           & Average real air temperature calculated from the sonic anemometer  \\
time                               & Hour of the day                                                    \\
ts (mean)                          & Average sonic temperature                                          \\
u (mean)                           & Mean wind velocity                                                 \\
u*                                 & Friction velocity                                                  \\
uu                                 & Variance of the streamwise (horizontal) wind speed (u)             \\
uw                                 & Covariance between streamwise wind speed and vertical wind speed   \\
vv                                 & Variance of the cross-wind wind speed (v)                          \\
wm                                 & Methane flux                                                       \\
wq                                 & Water flux                                                         \\
ww                                 & Variance of the vertical wind speed (w)                            \\
year                               & Calendar year                                                      \\
ze                                 & Zenith angle\\           \bottomrule                                           
\end{tabular}%
}
\caption{Meaning of labels of all predictive features included in the F\textsubscript{25} feature set.}
\label{tab:feature_meanings}
\end{table*}

\begin{table*}[]
\centering
\resizebox{\textwidth}{!}{%
\begin{tabular}{@{}llrrrrrrr@{}}
\toprule
\multicolumn{1}{c}{\textbf{Model}} &
  \multicolumn{1}{c}{\textbf{Feature Set}} &
  \multicolumn{1}{c}{\textbf{Adj. R2 Night}} &
  \multicolumn{1}{c}{\textbf{R2 Night}} &
  \multicolumn{1}{c}{\textbf{Adj. R2 Winter}} &
  \multicolumn{1}{c}{\textbf{R2 Winter}} &
  \multicolumn{1}{c}{\textbf{Adj. R2 Flood}} &
  \multicolumn{1}{c}{\textbf{R2 Flood}} &
  \multicolumn{1}{c}{\textbf{Time (total)}} \\ \midrule
XGBRegressor              & F\textsubscript{E}   & 0.94 & 0.94 & -0.25 & -0.25 & 0.60  & 0.60  & 22.81                \\
LGBMRegressor             & F\textsubscript{E}   & 0.94 & 0.94 & 0.03  & 0.03  & 0.67  & 0.67  & 2.22                 \\
GradientBoostingRegressor & F\textsubscript{E}   & 0.90 & 0.90 & -0.10 & -0.10 & 0.54  & 0.54  & 90.53                \\
DecisionTreeRegressor     & F\textsubscript{E}   & 0.89 & 0.89 & -0.76 & -0.76 & 0.23  & 0.23  & 4.35                 \\
KNNRegressor              & F\textsubscript{E}   & 0.79 & 0.79 & -2.10 & -2.10 & -1.43 & -1.42 & 2.43                 \\
LinearRegression          & F\textsubscript{E}   & 0.76 & 0.76 & -0.11 & -0.11 & -0.42 & -0.42 & 0.10                 \\
Ridge                     & F\textsubscript{E}   & 0.76 & 0.76 & -0.11 & -0.11 & -0.42 & -0.42 & 0.09                 \\
Neural Network            & F\textsubscript{E}   & 0.89 & 0.89 & 0.45  & 0.45  & 0.51  & 0.51  & \multicolumn{1}{l}{} \\
XGBRegressor              & F\textsubscript{25}  & 0.96 & 0.96 & 0.13  & 0.13  & 0.71  & 0.71  & 52.30                \\
LGBMRegressor             & F\textsubscript{25}  & 0.95 & 0.95 & 0.16  & 0.16  & 0.73  & 0.73  & 4.66                 \\
GradientBoostingRegressor & F\textsubscript{25}  & 0.92 & 0.92 & 0.34  & 0.34  & 0.65  & 0.65  & 337.38               \\
DecisionTreeRegressor     & F\textsubscript{25}  & 0.90 & 0.90 & -0.34 & -0.34 & 0.52  & 0.52  & 15.70                \\
KNNRegressor              & F\textsubscript{25}  & 0.85 & 0.85 & -1.84 & -1.83 & -1.33 & -1.31 & 166.73               \\
Ridge                     & F\textsubscript{25}  & 0.80 & 0.80 & 0.18  & 0.18  & 0.22  & 0.23  & 0.14                 \\
LinearRegression          & F\textsubscript{25}  & 0.80 & 0.80 & 0.18  & 0.18  & 0.22  & 0.23  & 0.20                 \\
XGBRegressor              & F\textsubscript{RFE} & 0.95 & 0.95 & 0.13  & 0.13  & 0.69  & 0.69  & 37.83                \\
LGBMRegressor             & F\textsubscript{RFE} & 0.95 & 0.95 & 0.14  & 0.14  & 0.73  & 0.73  & 3.46                 \\
GradientBoostingRegressor & F\textsubscript{RFE} & 0.92 & 0.92 & 0.37  & 0.37  & 0.67  & 0.67  & 3205.86              \\
DecisionTreeRegressor     & F\textsubscript{RFE} & 0.90 & 0.90 & -0.36 & -0.36 & 0.55  & 0.55  & 9.66                 \\
KNNRegressor              & F\textsubscript{RFE} & 0.84 & 0.84 & -1.83 & -1.83 & -1.33 & -1.32 & 150.55               \\ \bottomrule
\end{tabular}%
}
\caption{East End, R2 and Adjusted R2 results}
\label{tab:EE_Results_R2}
\end{table*}

\begin{table*}[]
\centering
\resizebox{\textwidth}{!}{%
\begin{tabular}{@{}llrrrrrrr@{}}
\toprule
\multicolumn{1}{c}{\textbf{Model}} &
  \multicolumn{1}{c}{\textbf{Feature Set}} &
  \multicolumn{1}{c}{\textbf{RMSE Night}} &
  \multicolumn{1}{c}{\textbf{Slope Night}} &
  \multicolumn{1}{c}{\textbf{RMSE Winter}} &
  \multicolumn{1}{c}{\textbf{Slope Winter}} &
  \multicolumn{1}{c}{\textbf{RMSE Flood}} &
  \multicolumn{1}{c}{\textbf{Slope Flood}} &
  \multicolumn{1}{c}{\textbf{Time (total)}} \\ \midrule
XGBRegressor              & F\textsubscript{E}   & 0.18 & 0.94 & 0.74 & 0.58  & 0.83 & 0.66 & 22.81                \\
LGBMRegressor             & F\textsubscript{E}   & 0.19 & 0.93 & 0.66 & 0.58  & 0.76 & 0.67 & 2.22                 \\
GradientBoostingRegressor & F\textsubscript{E}   & 0.24 & 0.88 & 0.70 & 0.53  & 0.90 & 0.52 & 90.53                \\
DecisionTreeRegressor     & F\textsubscript{E}   & 0.25 & 0.94 & 0.88 & 0.65  & 1.16 & 0.64 & 4.35                 \\
KNNRegressor              & F\textsubscript{E}   & 0.35 & 0.79 & 1.17 & -0.05 & 2.06 & 0.03 & 2.43                 \\
LinearRegression          & F\textsubscript{E}   & 0.37 & 0.76 & 0.70 & 0.63  & 1.58 & 0.12 & 0.10                 \\
Ridge                     & F\textsubscript{E}   & 0.37 & 0.76 & 0.70 & 0.63  & 1.58 & 0.12 & 0.09                 \\
Neural Network            & F\textsubscript{E}   & 0.24 & 0.88 & 0.79 & 0.95  & 0.75 & 0.45 & \multicolumn{1}{l}{} \\
XGBRegressor              & F\textsubscript{25}  & 0.16 & 0.95 & 0.62 & 0.74  & 0.71 & 0.74 & 52.30                \\
LGBMRegressor             & F\textsubscript{25}  & 0.17 & 0.94 & 0.61 & 0.72  & 0.69 & 0.73 & 4.66                 \\
GradientBoostingRegressor & F\textsubscript{25}  & 0.22 & 0.89 & 0.54 & 0.70  & 0.78 & 0.60 & 337.38               \\
DecisionTreeRegressor     & F\textsubscript{25}  & 0.24 & 0.94 & 0.77 & 0.68  & 0.92 & 0.70 & 15.70                \\
KNNRegressor              & F\textsubscript{25}  & 0.29 & 0.84 & 1.12 & -0.05 & 2.01 & 0.08 & 166.73               \\
Ridge                     & F\textsubscript{25}  & 0.34 & 0.81 & 0.60 & 0.61  & 1.17 & 0.41 & 0.14                 \\
LinearRegression          & F\textsubscript{25}  & 0.34 & 0.81 & 0.60 & 0.62  & 1.17 & 0.41 & 0.20                 \\
XGBRegressor              & F\textsubscript{RFE} & 0.16 & 0.95 & 0.62 & 0.70  & 0.74 & 0.74 & 37.83                \\
LGBMRegressor             & F\textsubscript{RFE} & 0.17 & 0.94 & 0.62 & 0.70  & 0.69 & 0.75 & 3.46                 \\
GradientBoostingRegressor & F\textsubscript{RFE} & 0.22 & 0.89 & 0.53 & 0.69  & 0.76 & 0.60 & 3205.86              \\
DecisionTreeRegressor     & F\textsubscript{RFE} & 0.24 & 0.95 & 0.78 & 0.71  & 0.89 & 0.74 & 9.66                 \\
KNNRegressor              & F\textsubscript{RFE} & 0.30 & 0.83 & 1.12 & -0.05 & 2.02 & 0.08 & 150.55               \\ \bottomrule
\end{tabular}%
}
\caption{East End RMSE and Slope Results}
\label{tab:EE_Results_RMSE_m}
\end{table*}

\begin{table*}[]
\centering
\resizebox{\textwidth}{!}{%
\begin{tabular}{@{}llrrrrrrr@{}}
\toprule
\multicolumn{1}{c}{\textbf{Model}} &
  \multicolumn{1}{c}{\textbf{Feature Set}} &
  \multicolumn{1}{c}{\textbf{Adj. R2 Night}} &
  \multicolumn{1}{c}{\textbf{Adj. R2 Winter}} &
  \multicolumn{1}{c}{\textbf{Adj. R2 Flood}} &
  \multicolumn{1}{c}{\textbf{R2 Night}} &
  \multicolumn{1}{c}{\textbf{R2 Winter}} &
  \multicolumn{1}{c}{\textbf{R2 Flood}} &
  \multicolumn{1}{c}{\textbf{Total Time (s)}} \\ \midrule
XGBRegressor              & F\textsubscript{E}   & 0.92 & -0.15 & 0.51  & 0.92 & -0.15 & 0.52  & 32.3                 \\
LGBMRegressor             & F\textsubscript{E}   & 0.91 & 0.56  & 0.65  & 0.91 & 0.56  & 0.65  & 2.8                  \\
GradientBoostingRegressor & F\textsubscript{E}   & 0.86 & 0.49  & 0.63  & 0.86 & 0.49  & 0.63  & 133.5                \\
DecisionTreeRegressor     & F\textsubscript{E}   & 0.85 & -0.41 & 0.08  & 0.85 & -0.41 & 0.08  & 7.0                  \\
KNNRegressor              & F\textsubscript{E}   & 0.79 & -0.10 & 0.02  & 0.79 & -0.10 & 0.02  & 4.1                  \\
Ridge                     & F\textsubscript{E}   & 0.65 & 0.29  & 0.51  & 0.65 & 0.29  & 0.51  & 0.1                  \\
LinearRegression          & F\textsubscript{E}   & 0.65 & 0.28  & 0.51  & 0.65 & 0.28  & 0.51  & 0.1                  \\
Neural Network            & F\textsubscript{E}   & 0.89 & 0.43  & 0.56  & 0.89 & 0.43  & 0.56  & \multicolumn{1}{l}{} \\
XGBRegressor              & F\textsubscript{25}  & 0.95 & 0.25  & 0.63  & 0.95 & 0.25  & 0.64  & 75.7                 \\
LGBMRegressor             & F\textsubscript{25}  & 0.94 & 0.59  & 0.71  & 0.94 & 0.59  & 0.72  & 6.0                  \\
GradientBoostingRegressor & F\textsubscript{25}  & 0.91 & 0.63  & 0.74  & 0.91 & 0.63  & 0.74  & 508.4                \\
DecisionTreeRegressor     & F\textsubscript{25}  & 0.89 & -0.06 & 0.27  & 0.89 & -0.05 & 0.28  & 24.5                 \\
KNNRegressor              & F\textsubscript{25}  & 0.85 & -0.19 & -0.16 & 0.85 & -0.19 & -0.15 & 310.6                \\
LinearRegression          & F\textsubscript{25}  & 0.77 & 0.52  & 0.52  & 0.77 & 0.52  & 0.53  & 0.3                  \\
Ridge                     & F\textsubscript{25}  & 0.77 & 0.52  & 0.52  & 0.77 & 0.52  & 0.53  & 0.2                  \\
XGBRegressor              & F\textsubscript{RFE} & 0.95 & -0.12 & 0.55  & 0.95 & -0.11 & 0.55  & 65.9                 \\
LGBMRegressor             & F\textsubscript{RFE} & 0.94 & 0.56  & 0.68  & 0.94 & 0.56  & 0.69  & 7.4                  \\
GradientBoostingRegressor & F\textsubscript{RFE} & 0.91 & 0.62  & 0.74  & 0.91 & 0.62  & 0.74  & 1126.0               \\
DecisionTreeRegressor     & F\textsubscript{RFE} & 0.89 & -0.20 & 0.28  & 0.89 & -0.20 & 0.28  & 15.2                 \\
KNNRegressor              & F\textsubscript{RFE} & 0.84 & -0.18 & -0.16 & 0.84 & -0.18 & -0.16 & 683.1                \\ \bottomrule
\end{tabular}%
}
\caption{Mayberry All Results - R2 and Adjusted R2}
\label{tab:MB_Results_R2}
\end{table*}

\begin{table*}[]
\centering
\resizebox{\textwidth}{!}{%
\begin{tabular}{@{}llrrrrrrr@{}}
\toprule
\multicolumn{1}{c}{\textbf{Model}} &
  \multicolumn{1}{c}{\textbf{Feature Set}} &
  \multicolumn{1}{c}{\textbf{RMSE Night}} &
  \multicolumn{1}{c}{\textbf{RMSE Winter}} &
  \multicolumn{1}{c}{\textbf{RMSE Flood}} &
  \multicolumn{1}{c}{\textbf{Slope Night}} &
  \multicolumn{1}{c}{\textbf{Slope Winter}} &
  \multicolumn{1}{c}{\textbf{Slope Flood}} &
  \multicolumn{1}{c}{\textbf{Total Time (s)}} \\ \midrule
XGBRegressor              & F\textsubscript{E}   & 0.28 & 0.83 & 0.79 & 0.92 & 0.46 & 0.58 & 32.3                 \\
LGBMRegressor             & F\textsubscript{E}   & 0.30 & 0.51 & 0.67 & 0.90 & 0.59 & 0.58 & 2.8                  \\
GradientBoostingRegressor & F\textsubscript{E}   & 0.37 & 0.55 & 0.69 & 0.83 & 0.58 & 0.72 & 133.5                \\
DecisionTreeRegressor     & F\textsubscript{E}   & 0.38 & 0.92 & 1.09 & 0.93 & 0.68 & 0.58 & 7.0                  \\
KNNRegressor              & F\textsubscript{E}   & 0.46 & 0.81 & 1.12 & 0.79 & 0.29 & 0.24 & 4.1                  \\
Ridge                     & F\textsubscript{E}   & 0.59 & 0.66 & 0.80 & 0.65 & 0.86 & 0.45 & 0.1                  \\
LinearRegression          & F\textsubscript{E}   & 0.59 & 0.66 & 0.80 & 0.65 & 0.86 & 0.45 & 0.1                  \\
Neural Network            & F\textsubscript{E}   & 0.24 & 0.60 & 0.57 & 0.88 & 0.59 & 0.48 & \multicolumn{1}{l}{} \\
XGBRegressor              & F\textsubscript{25}  & 0.23 & 0.67 & 0.68 & 0.95 & 0.65 & 0.75 & 75.7                 \\
LGBMRegressor             & F\textsubscript{25}  & 0.24 & 0.50 & 0.60 & 0.93 & 0.66 & 0.71 & 6.0                  \\
GradientBoostingRegressor & F\textsubscript{25}  & 0.30 & 0.47 & 0.58 & 0.88 & 0.69 & 0.72 & 508.4                \\
DecisionTreeRegressor     & F\textsubscript{25}  & 0.34 & 0.79 & 0.96 & 0.94 & 0.59 & 0.68 & 24.5                 \\
KNNRegressor              & F\textsubscript{25}  & 0.39 & 0.84 & 1.22 & 0.86 & 0.25 & 0.18 & 310.6                \\
LinearRegression          & F\textsubscript{25}  & 0.48 & 0.54 & 0.78 & 0.78 & 0.72 & 0.54 & 0.3                  \\
Ridge                     & F\textsubscript{25}  & 0.48 & 0.54 & 0.78 & 0.78 & 0.72 & 0.54 & 0.2                  \\
XGBRegressor              & F\textsubscript{RFE} & 0.23 & 0.81 & 0.76 & 0.95 & 0.54 & 0.69 & 65.9                 \\
LGBMRegressor             & F\textsubscript{RFE} & 0.24 & 0.52 & 0.64 & 0.93 & 0.64 & 0.69 & 7.4                  \\
GradientBoostingRegressor & F\textsubscript{RFE} & 0.31 & 0.48 & 0.58 & 0.88 & 0.68 & 0.71 & 1126.0               \\
DecisionTreeRegressor     & F\textsubscript{RFE} & 0.33 & 0.85 & 0.96 & 0.94 & 0.58 & 0.69 & 15.2                 \\
KNNRegressor              & F\textsubscript{RFE} & 0.40 & 0.84 & 1.22 & 0.85 & 0.25 & 0.18 & 683.1                \\ \bottomrule
\end{tabular}%
}
\caption{Mayberry All Results - RMSE and Slope}
\label{tab:MB_Results_RMSE_m}
\end{table*}

\begin{table*}[]
\centering
\resizebox{\textwidth}{!}{%
\begin{tabular}{@{}llrrrrrrr@{}}
\toprule
\multicolumn{1}{c}{\textbf{Model}} &
  \multicolumn{1}{c}{\textbf{Feature Set}} &
  \multicolumn{1}{c}{\textbf{Adj. R2 Night}} &
  \multicolumn{1}{c}{\textbf{Adj. R2 Flood}} &
  \multicolumn{1}{c}{\textbf{Adj. R2 Winter}} &
  \multicolumn{1}{c}{\textbf{R2 Night}} &
  \multicolumn{1}{c}{\textbf{R2 Winter}} &
  \multicolumn{1}{c}{\textbf{R2 Flood}} &
  \multicolumn{1}{c}{\textbf{Time (total)}} \\ \midrule
LGBMRegressor             & F\textsubscript{E}   & 0.94 & 0.62  & 0.79  & 0.94 & 0.79  & 0.62  & 1.7                  \\
XGBRegressor              & F\textsubscript{E}   & 0.94 & 0.60  & 0.45  & 0.94 & 0.46  & 0.60  & 12.4                 \\
GradientBoostingRegressor & F\textsubscript{E}   & 0.90 & 0.68  & 0.83  & 0.90 & 0.83  & 0.68  & 41.7                 \\
DecisionTreeRegressor     & F\textsubscript{E}   & 0.85 & 0.44  & -0.32 & 0.85 & -0.32 & 0.44  & 2.0                  \\
Ridge                     & F\textsubscript{E}   & 0.83 & 0.81  & 0.72  & 0.83 & 0.72  & 0.81  & 0.1                  \\
LinearRegression          & F\textsubscript{E}   & 0.83 & 0.81  & 0.72  & 0.83 & 0.72  & 0.81  & 0.1                  \\
KNNRegressor              & F\textsubscript{E}   & 0.80 & -0.33 & -0.68 & 0.80 & -0.68 & -0.32 & 1.7                  \\
Neural Network            & F\textsubscript{E}   & 0.89 & 0.53  & 0.69  & 0.89 & 0.69  & 0.53  & \multicolumn{1}{l}{} \\
XGBRegressor              & F\textsubscript{25}  & 0.95 & 0.31  & 0.70  & 0.95 & 0.70  & 0.31  & 26.3                 \\
LGBMRegressor             & F\textsubscript{25}  & 0.95 & 0.41  & 0.80  & 0.95 & 0.80  & 0.41  & 3.1                  \\
GradientBoostingRegressor & F\textsubscript{25}  & 0.92 & 0.32  & 0.84  & 0.92 & 0.84  & 0.32  & 157.8                \\
LinearRegression          & F\textsubscript{25}  & 0.88 & 0.82  & 0.64  & 0.88 & 0.65  & 0.82  & 0.1                  \\
Ridge                     & F\textsubscript{25}  & 0.88 & 0.83  & 0.62  & 0.88 & 0.63  & 0.83  & 0.1                  \\
DecisionTreeRegressor     & F\textsubscript{25}  & 0.86 & 0.24  & 0.34  & 0.86 & 0.34  & 0.24  & 6.8                  \\
KNNRegressor              & F\textsubscript{25}  & 0.84 & -0.75 & -0.27 & 0.84 & -0.26 & -0.74 & 61.4                 \\
LGBMRegressor             & F\textsubscript{RFE} & 0.95 & 0.41  & 0.80  & 0.95 & 0.80  & 0.41  & 2.3                  \\
XGBRegressor              & F\textsubscript{RFE} & 0.95 & 0.35  & 0.65  & 0.95 & 0.66  & 0.36  & 18.7                 \\
GradientBoostingRegressor & F\textsubscript{RFE} & 0.92 & 0.32  & 0.84  & 0.92 & 0.84  & 0.32  & 98.6                 \\
DecisionTreeRegressor     & F\textsubscript{RFE} & 0.86 & 0.30  & 0.36  & 0.86 & 0.36  & 0.31  & 4.2                  \\
KNNRegressor              & F\textsubscript{RFE} & 0.83 & -0.76 & -0.28 & 0.83 & -0.27 & -0.76 & 58.8                 \\ \bottomrule
\end{tabular}%
}
\caption{Sherman Wetland Results - R2 and Adjusted R2}
\label{tab:SW_Results_R2}
\end{table*}

\begin{table*}[]
\centering
\resizebox{\textwidth}{!}{%
\begin{tabular}{@{}llrrrrrrr@{}}
\toprule
\multicolumn{1}{c}{\textbf{Model}} &
  \multicolumn{1}{c}{\textbf{Feature Set}} &
  \multicolumn{1}{c}{\textbf{RMSE Night}} &
  \multicolumn{1}{c}{\textbf{RMSE Winter}} &
  \multicolumn{1}{c}{\textbf{RMSE Flood}} &
  \multicolumn{1}{c}{\textbf{Slope Night}} &
  \multicolumn{1}{c}{\textbf{Slope Winter}} &
  \multicolumn{1}{c}{\textbf{Slope Flood}} &
  \multicolumn{1}{c}{\textbf{Time (total)}} \\ \midrule
LGBMRegressor             & F\textsubscript{E}   & 0.28 & 0.40 & 1.41 & 0.93 & 1.01 & 0.58 & 1.7                  \\
XGBRegressor              & F\textsubscript{E}   & 0.28 & 0.64 & 1.46 & 0.94 & 0.95 & 0.63 & 12.4                 \\
GradientBoostingRegressor & F\textsubscript{E}   & 0.35 & 0.36 & 1.30 & 0.88 & 0.97 & 0.62 & 41.7                 \\
DecisionTreeRegressor     & F\textsubscript{E}   & 0.43 & 0.99 & 1.70 & 0.93 & 1.20 & 0.62 & 2.0                  \\
Ridge                     & F\textsubscript{E}   & 0.46 & 0.46 & 1.01 & 0.83 & 1.07 & 0.71 & 0.1                  \\
LinearRegression          & F\textsubscript{E}   & 0.46 & 0.46 & 1.01 & 0.83 & 1.07 & 0.71 & 0.1                  \\
KNNRegressor              & F\textsubscript{E}   & 0.50 & 1.12 & 2.64 & 0.81 & 0.08 & 0.09 & 1.7                  \\
Neural Network            & F\textsubscript{E}   & 0.24 & 0.54 & 0.56 & 0.88 & 0.87 & 0.43 & \multicolumn{1}{l}{} \\
XGBRegressor              & F\textsubscript{25}  & 0.25 & 0.47 & 1.90 & 0.95 & 0.92 & 0.39 & 26.3                 \\
LGBMRegressor             & F\textsubscript{25}  & 0.25 & 0.38 & 1.76 & 0.94 & 0.95 & 0.41 & 3.1                  \\
GradientBoostingRegressor & F\textsubscript{25}  & 0.32 & 0.34 & 1.88 & 0.90 & 0.92 & 0.36 & 157.8                \\
LinearRegression          & F\textsubscript{25}  & 0.39 & 0.52 & 0.97 & 0.88 & 1.11 & 0.70 & 0.1                  \\
Ridge                     & F\textsubscript{25}  & 0.39 & 0.53 & 0.95 & 0.88 & 1.12 & 0.71 & 0.1                  \\
DecisionTreeRegressor     & F\textsubscript{25}  & 0.42 & 0.70 & 1.99 & 0.92 & 0.98 & 0.37 & 6.8                  \\
KNNRegressor              & F\textsubscript{25}  & 0.45 & 0.97 & 3.02 & 0.85 & 0.22 & 0.02 & 61.4                 \\
LGBMRegressor             & F\textsubscript{RFE} & 0.26 & 0.39 & 1.76 & 0.94 & 0.93 & 0.42 & 2.3                  \\
XGBRegressor              & F\textsubscript{RFE} & 0.26 & 0.50 & 1.83 & 0.95 & 0.92 & 0.44 & 18.7                 \\
GradientBoostingRegressor & F\textsubscript{RFE} & 0.32 & 0.34 & 1.89 & 0.90 & 0.93 & 0.36 & 98.6                 \\
DecisionTreeRegressor     & F\textsubscript{RFE} & 0.42 & 0.69 & 1.90 & 0.92 & 0.97 & 0.40 & 4.2                  \\
KNNRegressor              & F\textsubscript{RFE} & 0.45 & 0.98 & 3.04 & 0.84 & 0.21 & 0.02 & 58.8                 \\ \bottomrule
\end{tabular}%
}
\caption{Sherman Wetland Results - RMSE and Slope}
\label{tab:SW_Results_RMSE_m}
\end{table*}

\begin{table*}[]
\centering
\resizebox{\textwidth}{!}{%
\begin{tabular}{@{}llrrrrrrrrr@{}}
\toprule
\multicolumn{1}{c}{\textbf{Model}} &
  \multicolumn{1}{c}{\textbf{Feature Set}} &
  \multicolumn{1}{c}{\textbf{Adj. R2 Night}} &
  \multicolumn{1}{c}{\textbf{Adj. R2 Winter}} &
  \multicolumn{1}{c}{\textbf{R2 Night}} &
  \multicolumn{1}{c}{\textbf{R2 Winter}} &
  \multicolumn{1}{c}{\textbf{RMSE Night}} &
  \multicolumn{1}{c}{\textbf{RMSE Winter}} &
  \multicolumn{1}{c}{\textbf{Slope Night}} &
  \multicolumn{1}{c}{\textbf{Slope Winter}} &
  \multicolumn{1}{c}{\textbf{Time (total)}} \\ \midrule
XGBRegressor              & F\textsubscript{E}   & 0.92 & -1.06  & 0.92 & -1.05  & 0.09 & 1.10  & 0.92 & 0.16   & 26.9                 \\
LGBMRegressor             & F\textsubscript{E}   & 0.91 & -0.62  & 0.91 & -0.62  & 0.10 & 0.98  & 0.89 & 0.23   & 3.0                  \\
GradientBoostingRegressor & F\textsubscript{E}   & 0.86 & -0.89  & 0.86 & -0.89  & 0.12 & 1.06  & 0.83 & 0.17   & 119.2                \\
KNNRegressor              & F\textsubscript{E}   & 0.81 & -1.69  & 0.81 & -1.69  & 0.14 & 1.26  & 0.82 & 0.02   & 2.7                  \\
DecisionTreeRegressor     & F\textsubscript{E}   & 0.80 & -0.86  & 0.80 & -0.86  & 0.15 & 1.05  & 0.91 & 0.23   & 6.3                  \\
Ridge                     & F\textsubscript{E}   & 0.77 & 0.71   & 0.77 & 0.72   & 0.16 & 0.41  & 0.77 & 0.63   & 0.1                  \\
LinearRegression          & F\textsubscript{E}   & 0.77 & 0.72   & 0.77 & 0.72   & 0.16 & 0.41  & 0.77 & 0.63   & 0.1                  \\
Neural Network            & F\textsubscript{E}   & 0.89 & 0.36   & 0.89 & 0.36   & 0.24 & 0.21  & 0.88 & 0.17   & \multicolumn{1}{l}{} \\
XGBRegressor              & F\textsubscript{25}  & 0.93 & -0.79  & 0.93 & -0.78  & 0.09 & 1.03  & 0.93 & 0.18   & 67.3                 \\
LGBMRegressor             & F\textsubscript{25}  & 0.92 & -0.68  & 0.92 & -0.67  & 0.09 & 1.00  & 0.90 & 0.21   & 5.8                  \\
GradientBoostingRegressor & F\textsubscript{25}  & 0.87 & -0.98  & 0.87 & -0.98  & 0.12 & 1.08  & 0.83 & 0.15   & 452.4                \\
KNNRegressor              & F\textsubscript{25}  & 0.85 & -1.56  & 0.85 & -1.56  & 0.13 & 1.23  & 0.82 & 0.02   & 220.3                \\
DecisionTreeRegressor     & F\textsubscript{25}  & 0.82 & -0.78  & 0.82 & -0.78  & 0.14 & 1.03  & 0.91 & 0.23   & 22.3                 \\
LinearRegression          & F\textsubscript{25}  & 0.81 & -infty & 0.81 & -infty & 0.14 & infty & 0.81 & -infty & 0.2                  \\
Ridge                     & F\textsubscript{25}  & 0.81 & 0.68   & 0.81 & 0.68   & 0.14 & 0.43  & 0.81 & 0.61   & 0.2                  \\
XGBRegressor              & F\textsubscript{RFE} & 0.93 & -1.03  & 0.93 & -1.03  & 0.09 & 1.10  & 0.93 & 0.13   & 47.7                 \\
LGBMRegressor             & F\textsubscript{RFE} & 0.92 & -0.72  & 0.92 & -0.72  & 0.10 & 1.01  & 0.90 & 0.21   & 4.0                  \\
GradientBoostingRegressor & F\textsubscript{RFE} & 0.87 & -0.98  & 0.87 & -0.97  & 0.12 & 1.08  & 0.83 & 0.15   & 285.7                \\
KNNRegressor              & F\textsubscript{RFE} & 0.84 & -1.56  & 0.84 & -1.56  & 0.13 & 1.23  & 0.81 & 0.02   & 206.1                \\
DecisionTreeRegressor     & F\textsubscript{RFE} & 0.83 & -0.78  & 0.83 & -0.78  & 0.14 & 1.03  & 0.91 & 0.24   & 13.6                 \\ \bottomrule
\end{tabular}%
}
\caption{West Pond Results - All}
\label{tab:WP_results_appendix}
\end{table*}

\end{document}